\documentclass[10pt,a4paper,conference]{IEEEtran}
\usepackage[utf8]{inputenc}
\usepackage{cite}
\usepackage[pdftex]{graphicx}  
\usepackage{soul}
\newcommand{\etal} {\textit{et~al.}}

\begin{document}

\title{ContourCNN:\\ convolutional neural network for contour data classification}

\author{\IEEEauthorblockN{Ahmad Droby}
\IEEEauthorblockA{Department of Computer Science\\
Ben Gurion University of the Negev\\
Email: drobya@post.bgu.ac.il}
\and
\IEEEauthorblockN{Jihad El-Sana}
\IEEEauthorblockA{Department of Computer Science\\
Ben Gurion University of the Negev\\
Email: el-sana@cs.bgu.ac.il}}

\maketitle

\begin{abstract}
This paper proposes a novel Convolutional Neural Network model for contour data analysis (ContourCNN) and shape classification. A contour is a circular sequence of points representing a closed shape. For handling the cyclical property of the contour representation, we employ circular convolution layers. Contours are often represented sparsely. To address information sparsity, we introduce priority pooling layers that select features based on their magnitudes. Priority pooling layers pool features with low magnitudes while leaving the rest unchanged. We evaluated the proposed model using letters and digits shapes extracted from the EMNIST dataset and obtained a high classification accuracy.
\end{abstract}


\IEEEpeerreviewmaketitle

\section{Introduction}
\label{sec:introduction}

In many computer vision tasks such as object detection, classification, and instance segmentation, object's shape offers vital information for the task at hand.

This paper focuses on 2D objects, which often represented using its bounding contour. Bounding contours are commonly represented as circular sequences of points or polygons. Such a representation plays an important role in many shape analysis tasks such as shape matching, classification, and segmentation.
However, a significant challenge of representing shapes using contour is the lack of common space, i.e. a shape can be represented by a huge number of contour representations, since the origin point can be defined anywhere on the contour of the shape, and the contour can have an arbitrarily sparsity level of points on the shape's contour.

Over the years a number of shape analysis methods were proposed that make the use of a designed smart features such as shape context for matching~\cite{belongie2002shape, ling2007shape}. Others such as~\cite{ramesh2015shape, bai2014shape} combine local and global features using the Bag of Words model for shape classification. In addition, there has been a number methods that uses machine learning algorithms such as SVM~\cite{ramesh2015shape} and deep Bayesian networks~\cite{li2016learning} to classify shapes using their features.

Over the last decades, convolutional neural networks (CNNs) have proved their effectiveness in addressing complex tasks such as classification, detection, and segmentation. CNNs can learn and extract features as opposed to traditional learning free methods that rely on hand-crafted features. However, CNNs are geared toward regular representation, such as vectors and matrices. Due to the irregular representation of contours, CNNs have not been used in shape analysis tasks.

In this paper, we propose a new CNN for contour analysis. The proposed network handles the lack of common space of the contour data mentioned above. It includes a circular convolutional layers and a priority pooling layers.

The circular convolutional layers deal with the circularity of the contour representation. To deal with the varying sparsity level of the contours, we introduce a number of priority pooling layers inspired by the pooling layer used in MeshCNN~\cite{hanocka2019meshcnn}, in which only points with low feature magnitude are pooled. To show the effectiveness of the proposed network, we apply it to classify shapes of digits and letters extracted from the EMNIST dataset .

The remainder of this paper is organised as follows: In section~\ref{sec:related_works} we summarize a number of related work done in the area of shape classification. We describe the proposed model in section~\ref{sec:model}. Training datasets and data representation are discussed in section~\ref{sec:dataset}. In section~\ref{sec:experiments}, experimental results are presented and discussed. Finally, we conclude and discuss future works in section~\ref{sec:conclusion}

\section{Related work}
\label{sec:related_works}

Existing solutions to the problem of shape classification can be split into two categories, learning and non-learning based solutions. 
Huttenlocher~\etal~\cite{huttenlocher1993comparing} and Liu~\etal~\cite{liu2010fast} compare and match shapes by calculating the similarity between two shapes using the Hausdorff and the Chamfer distance function, respectively.
Other approaches extract shape descriptors such as the Shape Context~\cite{belongie2002shape}, Inner Distance Shape Context~\cite{ling2007shape} and Height Functions~\cite{wang2012shape} to match and classify shape objects. Felzenszwalb~\etal~\cite{felzenszwalb2009object}
represent shapes' contours as a hierarchical structure of its fragments. This representation is used to classify the shape objects by a dynamic programming based matching algorithm. The Bag of Words (BoW) model was also applied to pool local features and generate a global descriptors for shape matching~\cite{ramesh2015shape, bai2014shape}.

Several learning based approaches were proposed to tackle shape matching. Approaches such as~\cite{ramesh2015shape, wang2014bag, shen2016shape} pool shape features using the BoW model and classify them using a SVM. Rameshet~\etal~\cite{ramesh2015shape} uses the spectral magnitude of the log-polar transform as local feature and bi-grams from the spatial co-occurrence matrix as contextual features, while~\cite{wang2014bag} describes a shape as local shape context features of its fragments. Ramesh\etal~\cite{shen2016shape} combines contour and skeleton information into local features to describe a shape object. Li~\etal~\cite{li2016learning} uses a deep Bayesian network trained by Markov chain Monte Carlo (MCMC) based optimization to classify shape descriptors of height features introduces by~\cite{wang2012shape}.
\section{Model}
\label{sec:model}
We propose a novel convolutional neural network that includes a custom-built 1D convolutional and pooling layers. Where the input of this network is a sequence of points representing a one layer contour. We introduce a 1D circular convolution layer, and a number of priority pooling layers. 

\subsection{1D Circular convolutional layer}
A sequence of circular data (such as a contour) is roll invariant, i.e. the sequence can be shifted (in a circular manner) and still represent the same data. Regular convolutional layers assume a unique start and end points in the input data, which removes the circularity of the data sequence. Thus, we apply circular convolutional layer. 

\begin{figure}
    \centering
    \includegraphics[width=0.4\textwidth]{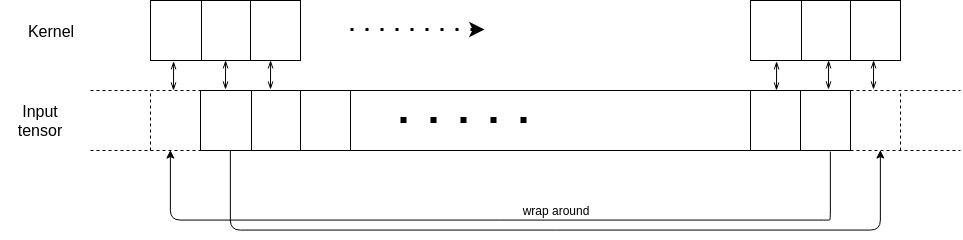}
    \caption{Circular convolution}
    \label{fig:cirular_conv}
\end{figure}

The circular convolution operation is defined as follow:
\begin{equation}
    u_{i} = (\sum_{j=0}^{M-1}{k_j\cdot x_{(i-\frac{M-1}{2} + j)\ mod\ N}}) + b_k
\end{equation}
where $x$ is the input 1D tensor with length $N$ and depth $D$, i.e. $x = [x_1, x_2, ..., x_N]$, where each $x_i$ is of size $1\times D$. $k$ is a  1D kernel of size $M$ and depth $D$ with bias $b_k$, and $u$ is the 1D output tensor of the convolution operation. As can be seen in Figure~\ref{fig:cirular_conv}, circular convolution work similar to regular convolution, but handles the end vertices differently. When the kernel is on a boundary of the input it wraps around to the other end. 

\subsection{Priority pooling layers}
\label{sec:priority_pooling}
The different vertices on a contour carry various levels of geometric information. A vertex that lies on a straight line which is defined by its immediate neighbours has little information; its position can be inferred from its two adjacent vertices.

\begin{figure}
    \centering
    \includegraphics[width=0.4\textwidth]{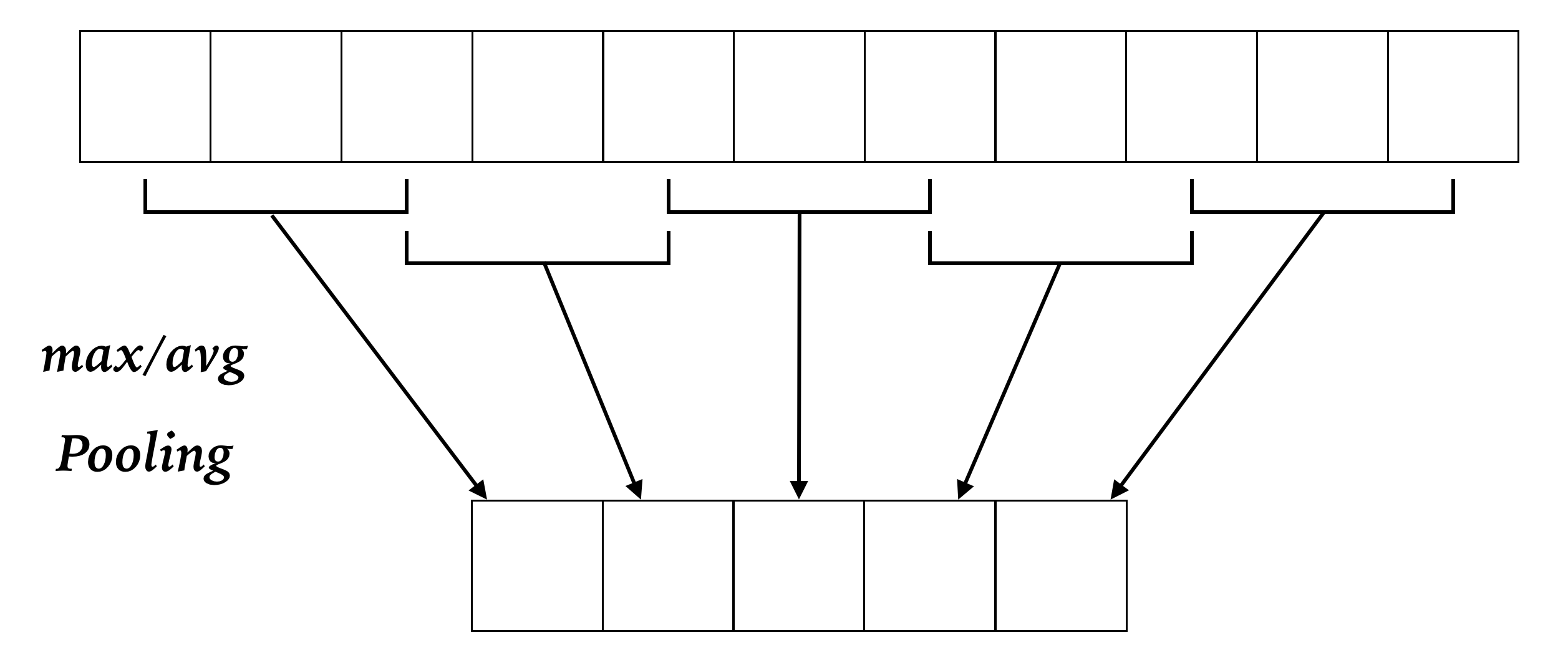}
    \caption{Illustration of the regular pooling operation with kernel of size three and stride value equal to two}
    \label{fig:regular_pooling}
\end{figure}

Typical pooling operations, such as average and max, are applied uniformally in a local manner, regardless of the global relative importance of the input vertices (as illustrated in Figure~\ref{fig:regular_pooling}). To overcome this limitation and apply irregular global pooling, we introduce priority pooling layers, that reduce the dimensions of an input tensor by iteratively considering the magnitude of the features.


\begin{figure}
    \centering
    \includegraphics[width=0.5\textwidth]{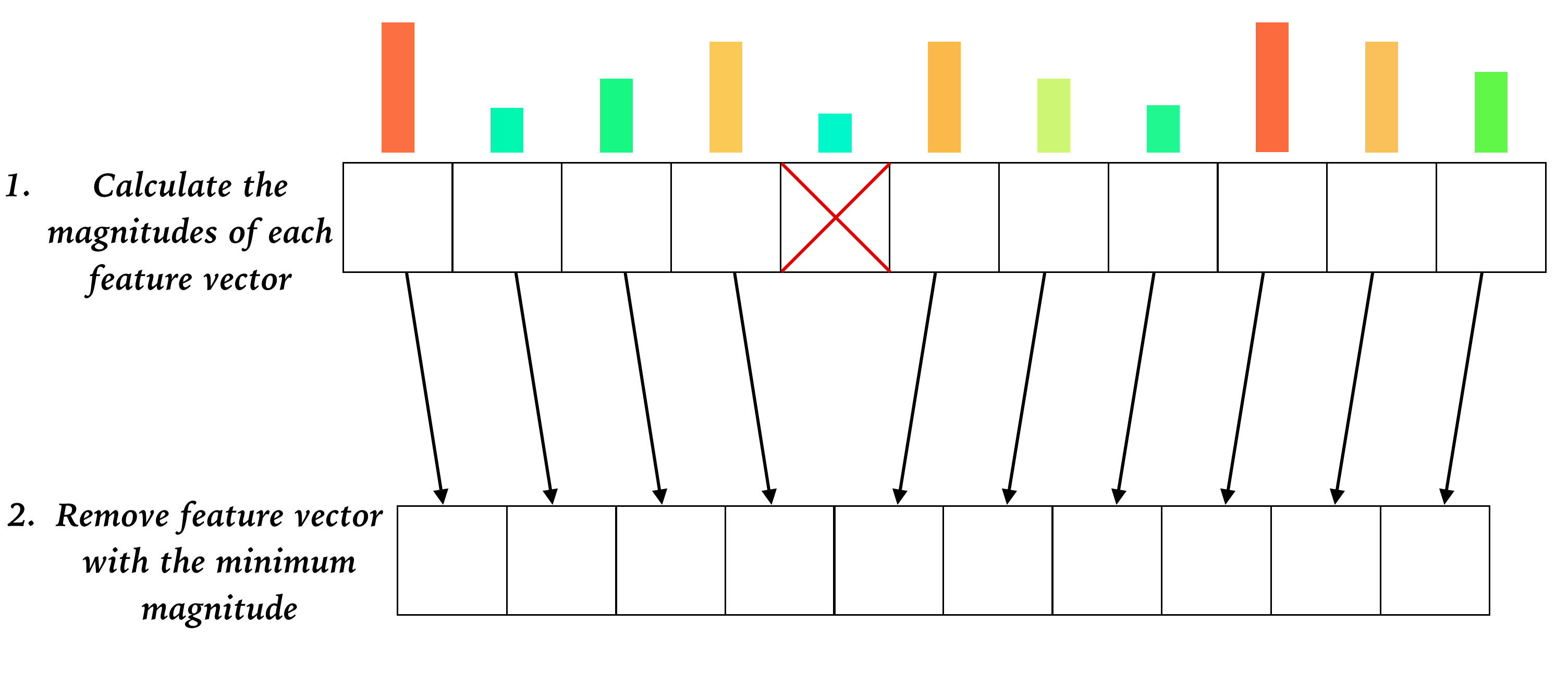}
    \caption{Illustration of the Remove One priority pooling operation. The colored bars represent the magnitude of each one of the feature vectors.}
    \label{fig:remove_one_priority_pooling}
\end{figure}

{\bf Remove One priority pooling} accepts an input tensor of dimensions $1\times N \times D$, where $N$ is the length of the input and $D$ is its depth (the number of features for each data point). The magnitude of each feature vertex is calculated using $L_2$ and utilized to prioritized the list of vertices, then the vertex with the lowest priority is removed (see Figure~\ref{fig:remove_one_priority_pooling}). This operation is repeated iteratively until the input tensor reaches a desired dimension. 
The desired dimensions we used are shown in Table~\ref{tab:contourcnn_network_conf}.

\begin{figure}
    \centering
    \includegraphics[width=0.5\textwidth]{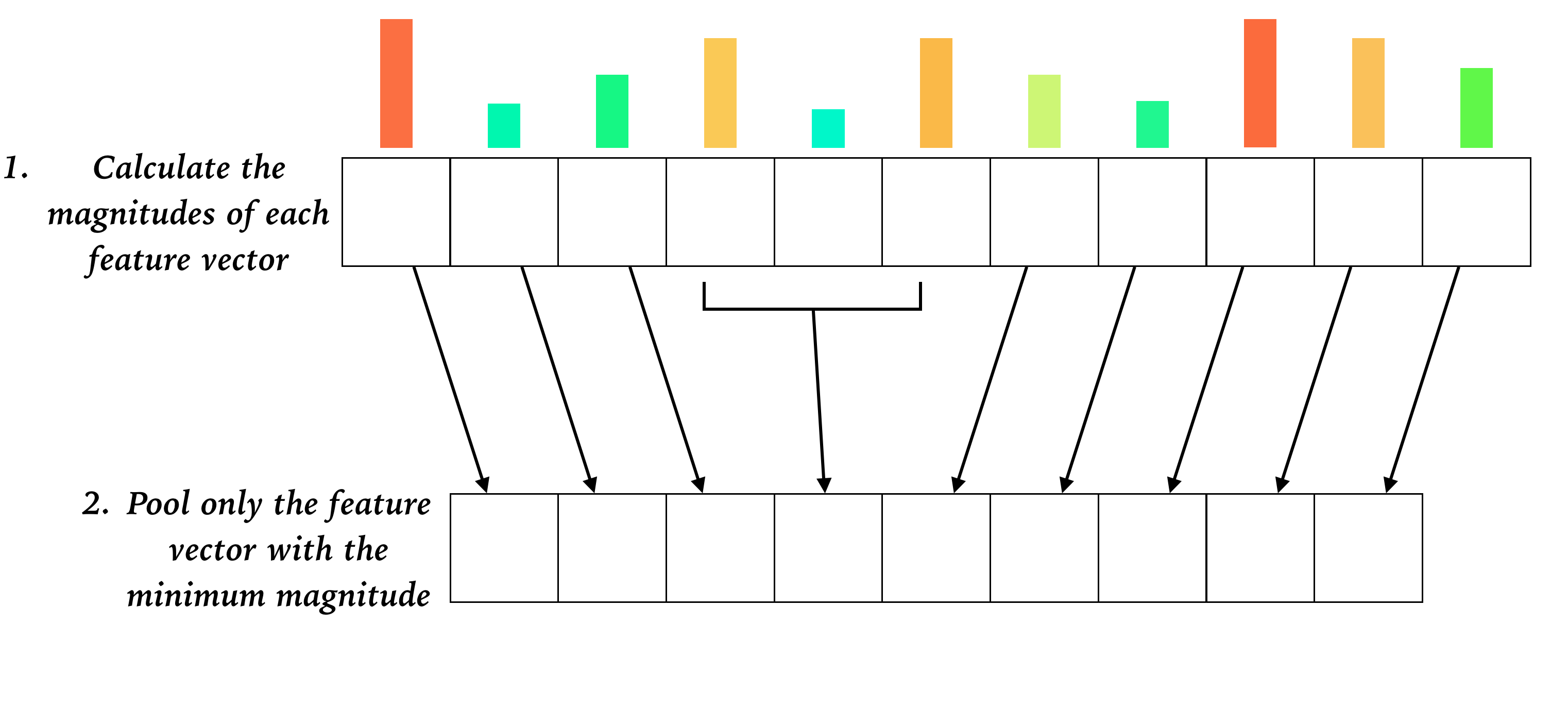}
    \caption{Illustration of the Max and Average priority pooling operation with kernel of size three. The colored bars represent the magnitude of each one of the feature vectors.}
    \label{fig:max_avg_priority_pooling}
\end{figure}

{\bf Max and Average priority pooling} layers work similar to the Remove One priority pooling. However, instead of removing the feature vertex with the lowest priority, the vertex with the lowest priority is pooled (either average or max). After each pooling operation the features' magnitudes of the remaining vertices are recalculated (see Figure~\ref{fig:remove_one_priority_pooling}). Similar to the Remove One, this priority pooling is applied iteratively until the input tensor reaches a desired dimension.

\subsection{Network architecture}
Similar to a typical 2D classification CNNs, our network architecture consists of two sub-networks: a feature extraction convolutional network, and a fully connected classification network.

The feature extraction network consists of four blocks of circular convolutional layers, a batch normalization layer, and a priority pooling layer. 
The extracted features are average pooled and fed to a fully connected classification network.\\
The architecture and configuration of the network can be seen in Table~\ref{tab:contourcnn_network_conf}. Such architecture were used to imitate AlexNet's and the parameters were further tuned through experiments.

\begin{table}[htb]
    \centering
    \begin{tabular}{c}
        \hline
         CircConv $f_{in} \times 32$\\
         PriorityPool $\rightarrow 40$\\
         CircConv $32 \times 64$\\
         PriorityPool $\rightarrow 30$\\
         CircConv $64 \times 128$\\
         PriorityPool $\rightarrow 20$\\
         GlobalAvgPool\\
         FCN $128 \times 80$\\
         FCN $80 \times f_{out}$\\
        \hline
        \\
    \end{tabular}
    \caption{Network configuration. CircConv $a \times b$ is a circular convolutional layer, where $a$ is the number of features (the depth) of the input, and $b$ is the output.  PriorityPool $\rightarrow N$ is a priority pooling layer, where $N$ is the length of the output. FCN $c\times d$ is a fully connected layer with $c$ input neurons and $d$ output neurons. $f_{in}$ is the number of features of the input of the network and $f_{out}$ is the number of outputs, i.e. number of classes}
    \label{tab:contourcnn_network_conf}
\end{table}


\section{Dataset}
\label{sec:dataset}

Existing shapes dataset such as MPEG-7~\cite{latecki2000shape} (1400 shapes of 70 classes) and Animal Shapes~\cite{bai2009integrating} (2000 shapes of 20 classes) contain insufficient number of samples to train CNNs models. In order to test the proposed model, we needed a sufficiently large dataset suitable for training deep-learning models. 
We chose to extract the contours from the EMNIST dataset~\cite{cohen2017emnist} which contain a large number of hand-written digits, uppercase and lowercase letter images. The contours were extracted using the algorithm~\cite{suzuki1985topological} implemented in the OpenCV library.\\
Our experiments were conducted on two separate subset of the EMNIST dataset. The subsets of digits consisting of images from 10 digit classes (0-9), a total of 402,953 images (344,307 for training and 58,646 for testing); and the subset of uppercase letters consisting of images from 26 classes (A-Z), a total of 220,304 images (208,364 for training and 11,941 for testing). We chose to split those two subset to avoid ambiguity between some of the letters and some of the digits, for example, in some cases the digits 1 and 0 cannot be differentiated from the letters I and O respectively.

\subsection{Data representation}
\label{sec:data_representation}
\begin{figure}
    \centering
    \includegraphics[width=0.5\textwidth]{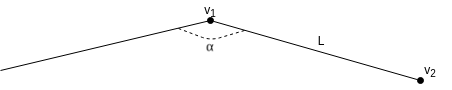}
    \caption{Polar representation of a contour. The vertex $v_1$ is represented by the angle $\alpha$ (the angle between the two lines), and the line length $L$ the length of the line between the vertex $v_1$ and the next vertex in the sequence $v_2$}
    \label{fig:contour_data_representation}
\end{figure}

We experimented with two different representation of the contour data. 
The first is a normalized Cartesian representation- each contour is represented as a sequence of the normalized $x,y \in [0,1]$ position points of its vertices.
In the second, each contour is represented as a sequence of an angles and a line lengths, we'll refer to this representation as the "Polar representation". Each angle and line length $(\alpha, L)$ represents a vertex on the contour. As shown in Figure~\ref{fig:contour_data_representation}, in each $(\alpha, L)$ vertex representation, $\alpha \in [-\pi, \pi]$ equals the angle between the two adjacent lines, and $L$ is the length of the line between the given vertex and the next vertex in the sequence.\\

\section{Experimental Evaluation}
\label{sec:experiments}
To evaluate the performance of our model, we conducted several experiments comparing activation functions, the different priority pooling layers, and the Cartesian and Polar representations (discussed in Section~\ref{sec:data_representation}). Furthermore, we compare the performance of our model with AlexNet. Those comparison were conducted on the letters subset, because it offer more complex and varied data than the digits subset. Finally, we report the results of the best performing model on the letter and the digits subsets from EMNIST. In each one of the experiments the model was trained and tested using the official train and test split of EMNIST.

\subsubsection{Activation function}
In this experiment we compared the performance of the model with different activation functions. We trained three networks with the activation functions, Sgimoid, TanH, and ReLU using the letters subset, the Cartesian representation, and with the Remove One priority pooling layers. 

\begin{figure}
    \centering
    \includegraphics[width=0.5\textwidth]{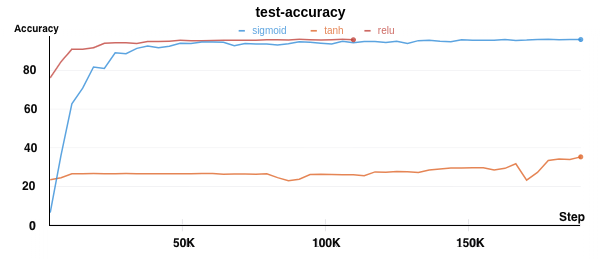}
    \includegraphics[width=0.5\textwidth]{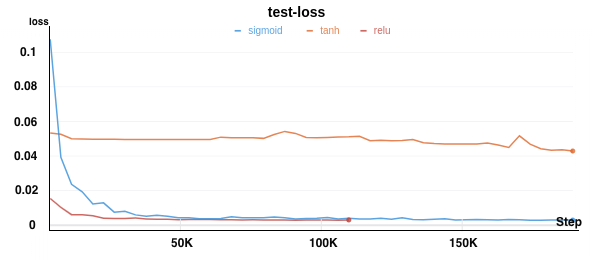}
    \caption{The loss and accuracy of a number of networks using different activation functions on the test set during the training process}
    \label{fig:contour_activation_results}
\end{figure}

Figure~\ref{fig:contour_activation_results} shows the test loss and accuracy of each network during the training process. We can see that the two networks using Sigmoid and ReLU activation function reached similar accuracy and loss values. Using the ReLU activation function the network reached a higher accuracy faster than using the Sigmoid function. While using TanH activation function the network was unable to learn and got stuck in a local minima.

\subsubsection{Priority pooling layers}
In this experiment we compare the different priority pooling layer introduced in Section~\ref{sec:priority_pooling}. 
Similar to the previous experiment, three networks with different priority pooling layers and ReLU activation function were trained using the letters subset and the Cartesian representation.
Note that this compression only affect the first three priority pooling layers, the global average pooling layer preceding the full connected layers remains unchanged.

\begin{figure}
    \centering
    \includegraphics[width=0.5\textwidth]{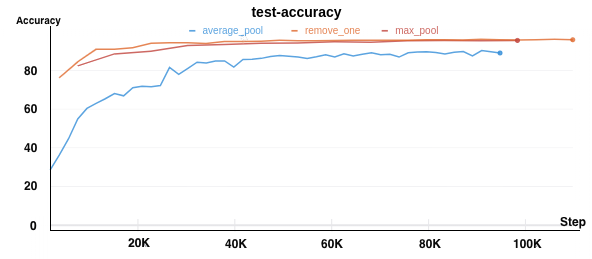}
    \includegraphics[width=0.5\textwidth]{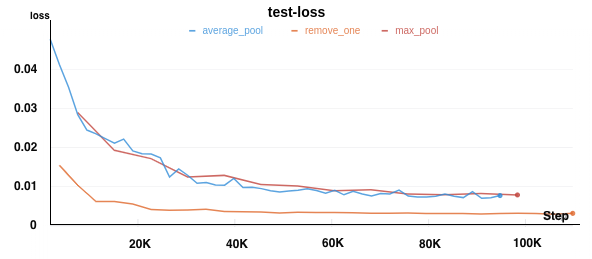}
    \caption{The loss and accuracy of networks using different priority pooling layers on the test set during the training process}
    \label{fig:contour_pooling_results}
\end{figure}

Figure~\ref{fig:contour_pooling_results} shows the accuracy and loss of the networks on the test subset. As can be seen the networks which include the Remove One and the Max priority pooling layers reached similar accuracy. However, the two networks achieve different loss values on the test set. This suggests, the network with the Remove One priority pooling is more confident concern its classification, i.e. it gives a higher probability to the correct classification. As seen, the networks with the average priority pooling layer converges and obtained around 90\% accuracy rate on the test subset, but it did not reach a higher accuracy as the other two networks.

\subsubsection{Experimenting with data representations}
In the following experiment we trained two identical networks with Remove One priority pooling layers and ReLU activation function on the letters subsets with the Cartesian and Polar data representations mentioned in Section~\ref{sec:data_representation}. 

\begin{figure}
    \centering
    \includegraphics[width=0.5\textwidth]{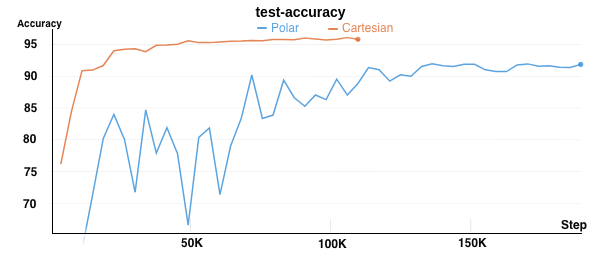}
    \caption{The accuracy of two networks on the test set during training. The two networks were trained using two different data representations, the Cartesian representation and the Polar representation}
    \label{fig:contour_data_representation_accuracy}
\end{figure}

\begin{figure}
    \centering
    \includegraphics[width=0.15\textwidth]{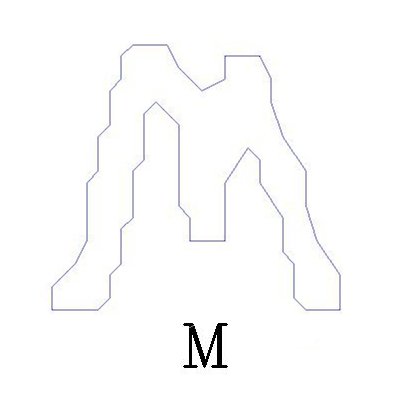}    \includegraphics[width=0.15\textwidth]{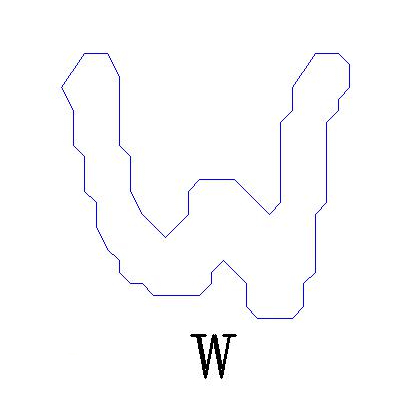}
    \includegraphics[width=0.15\textwidth]{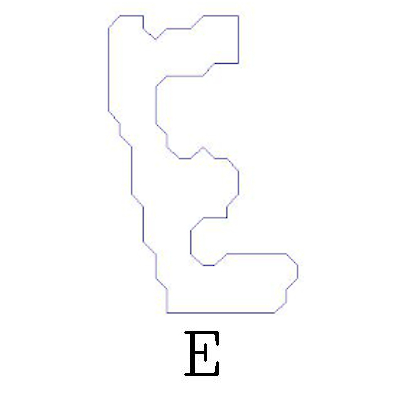}
    \caption{Example contour instances of the letters M, W and E in the test subset}
    \label{fig:contour_m_e_AL}
\end{figure}

\begin{figure}
    \centering
    \includegraphics[width=0.5\textwidth]{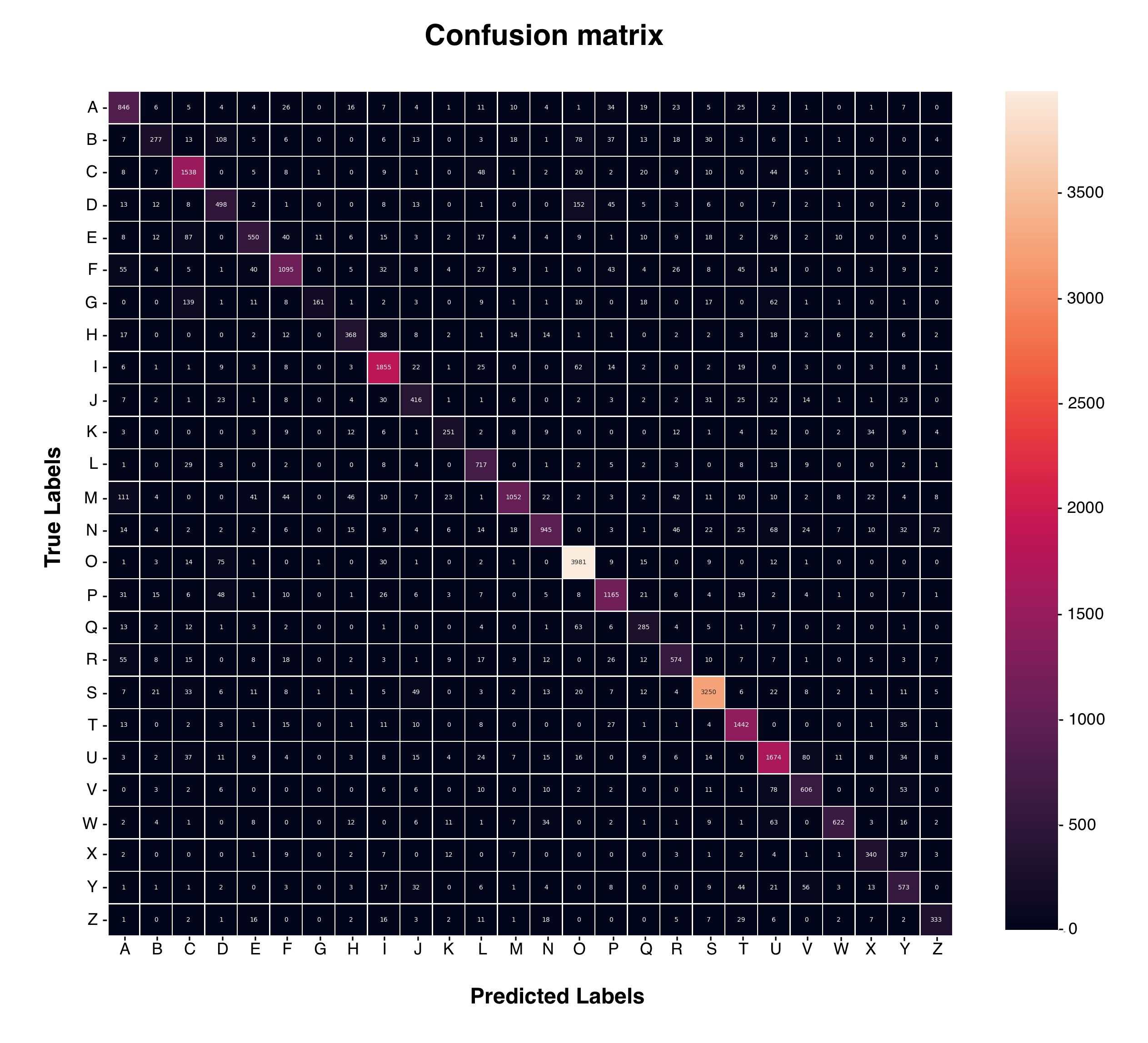}
    \caption{The confusion matrix of the network trained using the Polar representation. The color of each cell represent the number of samples classified in it. Lighter color means more samples (as shown in the color bar to the right)}
    \label{fig:contour_AL_confusion_matrix}
\end{figure}

Figure~\ref{fig:contour_data_representation_accuracy} shows the accuracy of the two networks on the test subset during training. Unsurprisingly, the network trained using the Cartesian representation reached higher accuracy faster than the network using the Polar representation. There will always be ambiguity between certain letters in the Polar representation. As can be seen in Figure~\ref{fig:contour_m_e_AL}, the letters M, W, and E have an almost identical Polar representation, since this representation is rotation invariant, rotating the letter M $180^{\circ}$ results in the letter W with the same Polar representation as the letter M. Such ambiguity can be seen in the confusion matrix of the model in Figure~\ref{fig:contour_AL_confusion_matrix}.

\begin{figure}
    \centering
    \includegraphics[width=0.11\textwidth]{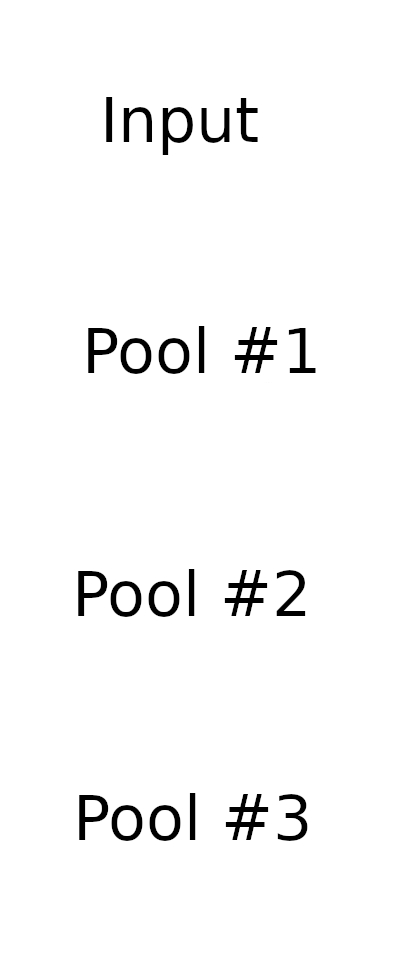}   
    \includegraphics[width=0.11\textwidth]{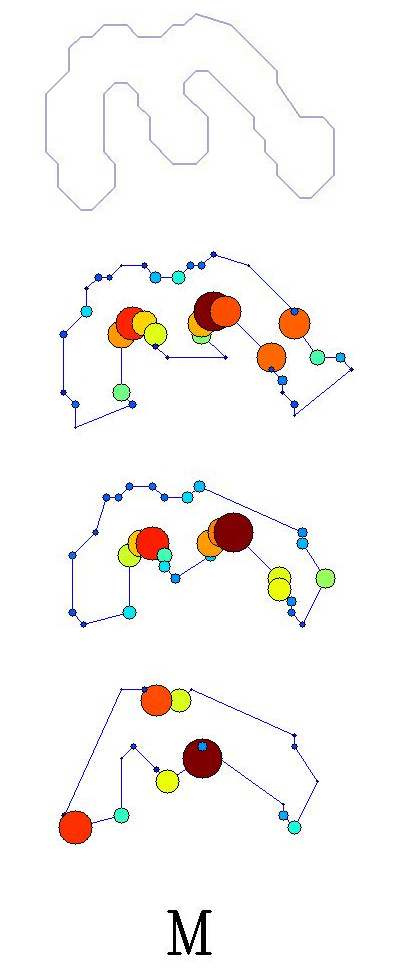}    \includegraphics[width=0.11\textwidth]{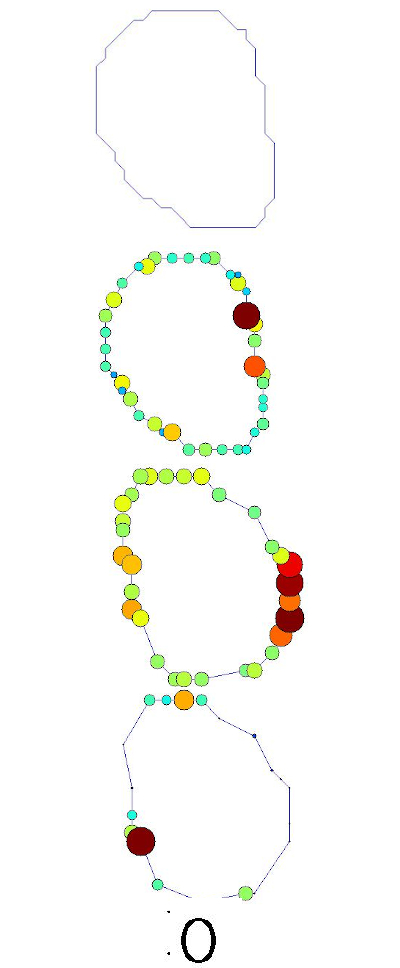}
    \includegraphics[width=0.11\textwidth]{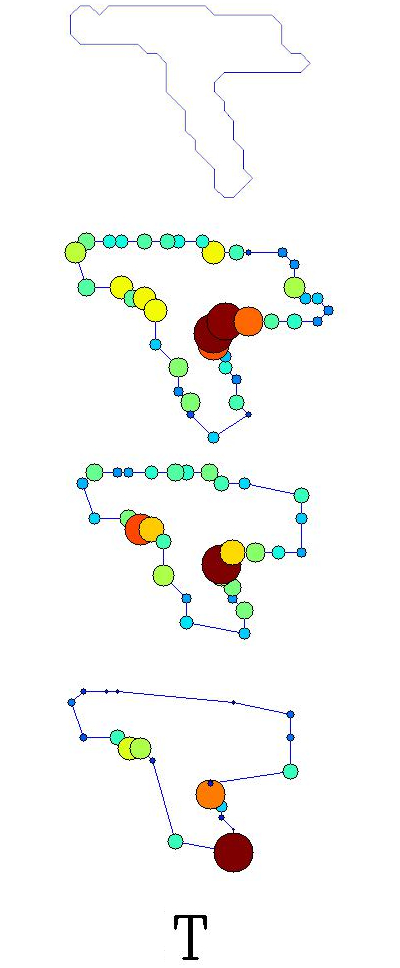}
    \caption{This figure show the simplification effect happening in the network. The row shows the input contour of the letter, the resulting contour after each pooling layer. The coloured circles represent the activations of each contour vertices; the size and colour of the circles reflect the magnitude of each activation}
    \label{fig:contour_letters_simplification}
\end{figure}

\paragraph{Shape simplification effects} the priority pooling layers pools/removes a selected features corresponding to points on input contour. Thus, we can visualize the progress of the network by redrawing the contour after each priority pooling without the pooled points. 
Interestingly, we observed that the network trained using the Polar representation seemingly simplifies the contour shape.
As can be seen in Figure~\ref{fig:contour_letters_simplification}, when drawing the contour after each priority pooling layer in the network, the contour of the letters is simplified as it traverse down the network's pooling layers.
This can be attributed to the fact that the activation of vertices that contain more information are larger than the activations of those with fewer information. Therefore, the priority pooling layers will first pool/remove redundant contour vertices or those with little information.

\subsubsection{Comparing with regular CNN classification model}
\begin{figure}
    \centering
    \includegraphics[width=0.15\textwidth]{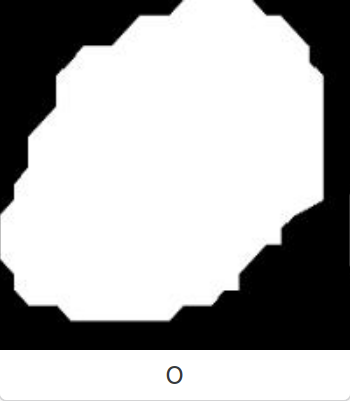}    \includegraphics[width=0.15\textwidth]{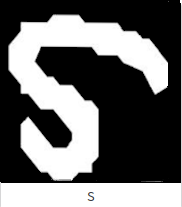}
    \includegraphics[width=0.15\textwidth]{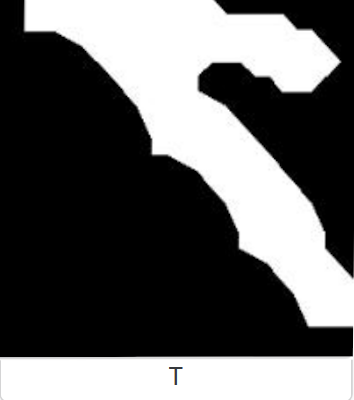}
    \caption{Example of the generated 2D images from the contour data of the letters O, S, and T}
    \label{fig:2d_generated_data}
\end{figure}

Here we compare our approach with a more established 2D classification CNN. 
Since the ContourCNN model are able to process one layer contours, training a 2D CNN using the original data would give it an unfair advantage. Because the original images include information missing from the contour data (e.g. the holes inside the letters A,B, D, and O). Therefore, in order to compare the two approaches, we generated 2D images by drawing the filled polygon represented by the extracted contour data (the silhouette of the letters) as can be seen in Figure~\ref{fig:2d_generated_data}. This way we insure that the two approaches are trained in a fair manner, i.e. using the same level of detail/information.
Since AlexNet~\cite{krizhevsky2012imagenet} has a similar architecture to our model, we chose to train it on the generated 2D images. The configuration of the network can be seen in Table~\ref{tab:alexnet_network_conf}.

\begin{figure}
    \centering
    \includegraphics[width=0.5\textwidth]{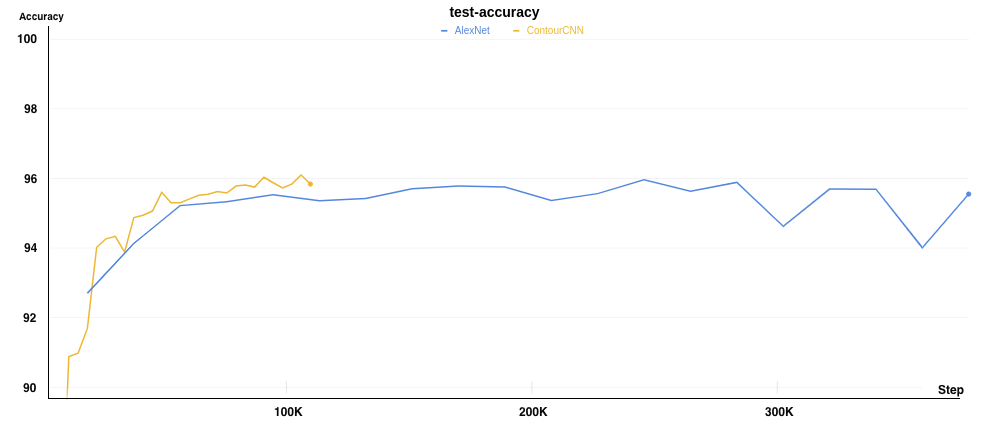}
    \caption{The test accuracy of two approaches; ContourCNN: training using contour data, AlexNet: CNN training using the generated 2D images.}
    \label{fig:contourcnn_alexnet_accuracy}
\end{figure}

As can be seen in Figure~\ref{fig:contourcnn_alexnet_accuracy} the two approaches reach similar accuracy of around 96\%. This results shows the effectiveness of the proposed model as it managed to achieve similar accuracy results to a more established model such as AlexNet. 

\begin{table}[htb]
    \centering
    \begin{tabular}{c}
        \hline
         Conv2D $f_{in} \times 64$\\
         MaxPool2D $k=3, s=2$\\
         Conv2D $64 \times 192$\\
         MaxPool2D $k=3, s=2$\\
         Conv2D $192 \times 384$\\
         Conv2D $384 \times 256$\\
         MaxPool2D $k=3, s=2$\\
         AvgPool $\rightarrow (6,6)$\\
         FCN $6\cdot 6 \cdot 256 \times 4096$\\
         FCN $4096 \times f_{out}$\\
        \hline
        \\
    \end{tabular}
    \caption{AlexNet configuration}
    \label{tab:alexnet_network_conf}
\end{table}

\subsubsection{Results of the best model on the letters and digits subsets}

\begin{figure}
    \centering
    \includegraphics[width=0.5\textwidth]{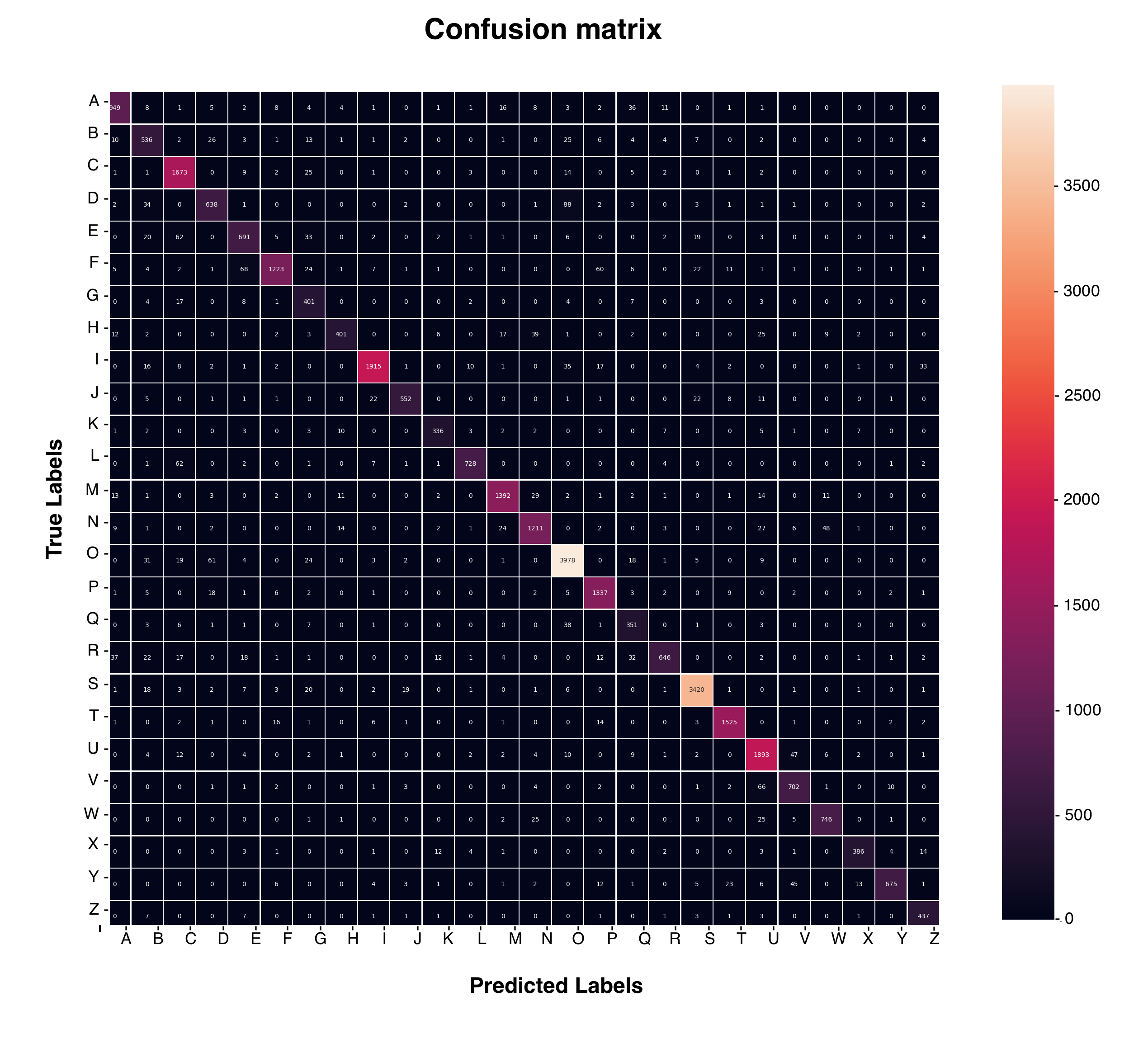}
    \caption{The confusion matrix of the classification results of the network trained on the letters subset. The color of each cell represent the number of samples classified in it. Lighter color means more samples (as shown in the color bar to the right)}
    \label{fig:contour_letters_confusion_matrix}
\end{figure}

\begin{figure}
    \centering
    \includegraphics[width=0.5\textwidth]{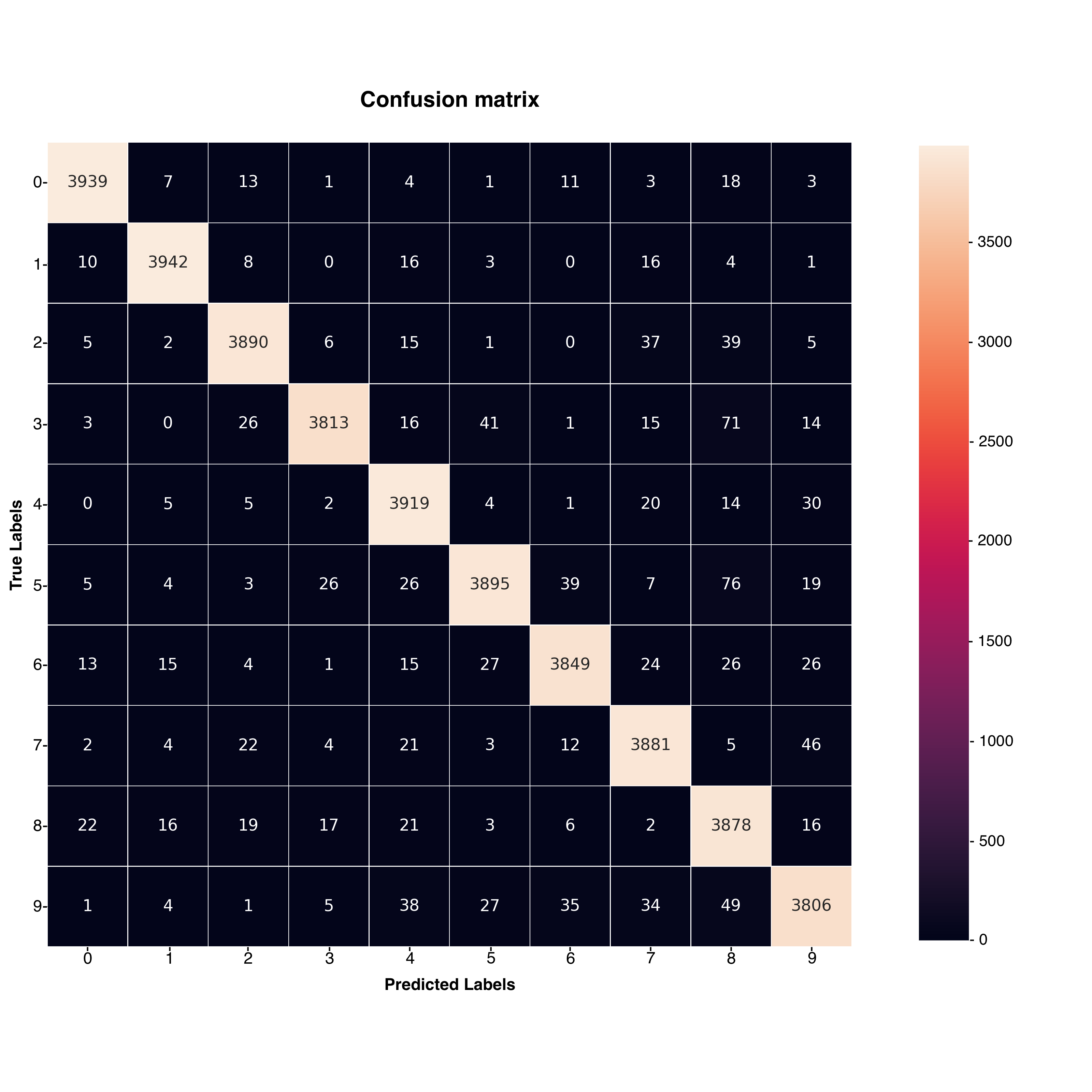}
    \caption{The confusion matrix of the classification results of the network trained on the digits subset.
    The color of a cell indicates the number of samples were used for classification.
    Lighter color means more samples (as shown in the color bar to the right)}
    \label{fig:contour_digits_confusion_matrix}
\end{figure}

The best results were reached by the model with Remove One priority pooling layer, ReLU activation function, and trained using the Cartesian representation. This model achieved around a 96\% accuracy rate on the uppercase letters subset and round a 96.7\% accuracy on the digits subset. The Figures~\ref{fig:contour_letters_confusion_matrix} and~\ref{fig:contour_digits_confusion_matrix} show the confusion matrices for the results of the trained network on the uppercase letters and digits subsets.

\section{Conclusion \& Future work}
\label{sec:conclusion}
This paper presents ContourCNN, a CNN for contour data analysis. We introduced 1D circular convectional layer for convolution on a circular data representation, which preserves the circularity of the data after the operation. To deal with the varying sparsity level of information of the contour data we introduced a number of priority pooling layers that pools only the vertices with the lowest feature magnitude. We showed the effectiveness of the model by classifying the contours of the digits and letters from the EMNIST dataset and compared the results to the similarly structured CNN, AlexNet. In addition, we provided a comparison of the accuracy and loss between the different priority pooling layers, activation function, and the input data representation.

In future work, we plan to explore the use of ContourCNN for contour segmentation using an encoder-decoder architecture of the model, where the pooled vertices are remembered in the down-sampling stage of the encoder and unpooled in the up-sampling stage of the decoder. We will also consider extending the ContourCNN to process nested contour. 

Lastly, we would like to end on the following note: while we used the proposed model to analyse contours, we realized it can be used to process other similar data forms (data with circular structure or data sequences that include redundant information).

\bibliography{M335}

\begin{thebibliography}{10}

\bibitem{belongie2002shape}
S.~Belongie, J.~Malik, and J.~Puzicha, ``Shape matching and object recognition
  using shape contexts,'' {\em IEEE Transactions on Pattern Analysis \& Machine
  Intelligence}, no.~4, pp.~509--522, 2002.

\bibitem{ling2007shape}
H.~Ling and D.~W. Jacobs, ``Shape classification using the inner-distance,''
  {\em IEEE transactions on pattern analysis and machine intelligence},
  vol.~29, no.~2, pp.~286--299, 2007.

\bibitem{ramesh2015shape}
B.~Ramesh, C.~Xiang, and T.~H. Lee, ``Shape classification using invariant
  features and contextual information in the bag-of-words model,'' {\em Pattern
  Recognition}, vol.~48, no.~3, pp.~894--906, 2015.

\bibitem{bai2014shape}
X.~Bai, C.~Rao, and X.~Wang, ``Shape vocabulary: A robust and efficient shape
  representation for shape matching,'' {\em IEEE Transactions on Image
  Processing}, vol.~23, no.~9, pp.~3935--3949, 2014.

\bibitem{li2016learning}
C.~Li, A.~Stevens, C.~Chen, Y.~Pu, Z.~Gan, and L.~Carin, ``Learning weight
  uncertainty with stochastic gradient mcmc for shape classification,'' in {\em
  Proceedings of the IEEE Conference on Computer Vision and Pattern
  Recognition}, pp.~5666--5675, 2016.

\bibitem{hanocka2019meshcnn}
R.~Hanocka, A.~Hertz, N.~Fish, R.~Giryes, S.~Fleishman, and D.~Cohen-Or,
  ``Meshcnn: a network with an edge,'' {\em ACM Transactions on Graphics
  (TOG)}, vol.~38, no.~4, p.~90, 2019.

\bibitem{huttenlocher1993comparing}
D.~P. Huttenlocher, G.~A. Klanderman, and W.~J. Rucklidge, ``Comparing images
  using the hausdorff distance,'' {\em IEEE Transactions on pattern analysis
  and machine intelligence}, vol.~15, no.~9, pp.~850--863, 1993.

\bibitem{liu2010fast}
M.-Y. Liu, O.~Tuzel, A.~Veeraraghavan, and R.~Chellappa, ``Fast directional
  chamfer matching,'' in {\em 2010 IEEE Computer Society Conference on Computer
  Vision and Pattern Recognition}, pp.~1696--1703, IEEE, 2010.

\bibitem{wang2012shape}
J.~Wang, X.~Bai, X.~You, W.~Liu, and L.~J. Latecki, ``Shape matching and
  classification using height functions,'' {\em Pattern Recognition Letters},
  vol.~33, no.~2, pp.~134--143, 2012.

\bibitem{felzenszwalb2009object}
P.~F. Felzenszwalb, R.~B. Girshick, D.~McAllester, and D.~Ramanan, ``Object
  detection with discriminatively trained part-based models,'' {\em IEEE
  transactions on pattern analysis and machine intelligence}, vol.~32, no.~9,
  pp.~1627--1645, 2009.

\bibitem{wang2014bag}
X.~Wang, B.~Feng, X.~Bai, W.~Liu, and L.~J. Latecki, ``Bag of contour fragments
  for robust shape classification,'' {\em Pattern Recognition}, vol.~47, no.~6,
  pp.~2116--2125, 2014.

\bibitem{shen2016shape}
W.~Shen, Y.~Jiang, W.~Gao, D.~Zeng, and X.~Wang, ``Shape recognition by bag of
  skeleton-associated contour parts,'' {\em Pattern Recognition Letters},
  vol.~83, pp.~321--329, 2016.

\bibitem{latecki2000shape}
L.~J. Latecki, R.~Lakamper, and T.~Eckhardt, ``Shape descriptors for non-rigid
  shapes with a single closed contour,'' in {\em Proceedings IEEE Conference on
  Computer Vision and Pattern Recognition. CVPR 2000 (Cat. No. PR00662)},
  vol.~1, pp.~424--429, IEEE, 2000.

\bibitem{bai2009integrating}
X.~Bai, W.~Liu, and Z.~Tu, ``Integrating contour and skeleton for shape
  classification,'' in {\em 2009 IEEE 12th international conference on computer
  vision workshops, ICCV workshops}, pp.~360--367, IEEE, 2009.

\bibitem{cohen2017emnist}
G.~Cohen, S.~Afshar, J.~Tapson, and A.~van Schaik, ``Emnist: an extension of
  mnist to handwritten letters,'' {\em arXiv preprint arXiv:1702.05373}, 2017.

\bibitem{suzuki1985topological}
S.~Suzuki {\em et~al.}, ``Topological structural analysis of digitized binary
  images by border following,'' {\em Computer vision, graphics, and image
  processing}, vol.~30, no.~1, pp.~32--46, 1985.

\bibitem{krizhevsky2012imagenet}
A.~Krizhevsky, I.~Sutskever, and G.~E. Hinton, ``Imagenet classification with
  deep convolutional neural networks,'' in {\em Advances in neural information
  processing systems}, pp.~1097--1105, 2012.

\end{thebibliography}
\bibliographystyle{ieeetr}

\end{document}